\def\year{2020}\relax
\documentclass[letterpaper]{article} 
\usepackage{aaai20}  
\usepackage{times}  
\usepackage{helvet} 
\usepackage{courier}  
\usepackage[hyphens]{url}  
\usepackage{graphicx} 
\usepackage{graphicx}  
\usepackage{amsmath}
\title{SESF-Fuse: An Unsupervised Deep Model for Multi-Focus Image Fusion}

\author{Boyuan Ma, \textsuperscript{\rm 1,2,3} Xiaojuan Ban,  \textsuperscript{\rm 1,2,3}\thanks{~Corresponding authors: banxj@ustb.edu.cn; huanghy@mater.ustb.edu.cn.} Haiyou Huang, \textsuperscript{\rm 1,4$*$} Yu Zhu\textsuperscript{\rm 1,2,3}\\
\textsuperscript{\rm 1}Beijing Advanced Innovation Center for Materials Genome Engineering, University of Science and Technology Beijing, China.\\ 
\textsuperscript{\rm 2}School of Computer and Communication Engineering, University of Science and Technology Beijing, China.\\
\textsuperscript{\rm 3}Beijing Key Laboratory of Knowledge Engineering for Materials Science, Beijing, China.\\
\textsuperscript{\rm 4}Institute for Advanced Materials and Technology, University of Science and Technology Beijing, China.\\
}

\begin{document}

\maketitle

\begin{abstract}
In this work, we propose a novel unsupervised deep learning model to address multi-focus image fusion problem. First, we train an encoder-decoder network in unsupervised manner to acquire deep feature of input images. And then we utilize these features and spatial frequency to measure activity level and decision map. Finally, we apply some consistency verification methods to adjust the decision map and draw out fused result. The key point behind of proposed method is that only the objects within the depth-of-field (DOF) have sharp appearance in the photograph while other objects are likely to be blurred. In contrast to previous works, our method analyzes sharp appearance in deep feature instead of original image. Experimental results demonstrate that the proposed method achieves the state-of-art fusion performance compared to existing 16 fusion methods in objective and subjective assessment. 
\end{abstract}

\section{Introduction}
\label{sec:introduction}

\noindent 
In recent years, multi-focus image fusion has become an important issue in image processing field. Due to the limited DOF of optical lenses, it is difficult to have all objects with quite different distances from the camera to be all-in-focus within one shot~\cite{LI2017100}. Therefore many researchers devoted to designing algorithm to fuse multiple images of the same scene but with different focus points to create an all-in-focus fused image. The fused image can be used for human or computer operators, and for further image-processing tasks such as segmentation, feature extraction and object recognition.

With the unprecedented success of deep learning, many fusion methods based on deep learning have been proposed. ~\cite{LIU2017191} first presented a CNN-based fusion method for multi-focus image fusion task. They used gaussian filter to generate synthetic images with different blurred levels to train a two-class image classification network. By using such supervised learning strategy, the network could distinguish whether the patch is in focus. After that, DeepFuse~\cite{Prabhakar_2017_ICCV} has been developed in an unsupervised manner to fuse multi-exposure images. DenseFuse~\cite{Li_2019_TIP} has been designed to fuse infrared and visible images, it utilized unsupervised encoder-decoder network to extract deep features of images and designed L1-norm fusion strategy to fuse two feature maps, and then, the decoder used fused features to obtain a fused image. The basic assumption behind this approach is that the L1 norm of feature vector for each node represents activity level of that. It can be applied to infrared and visible image fusion task. But for multi-focus task, it is commonly assumed that only the objects within the DOF have sharp appearance in the photograph while other objects are likely to be blurred~\cite{LIU2017191}. Therefore, we assume that in multi-focus task, what really matter is feature gradient, not feature intensity.

In order to verify this assumption, we present a fusion method based on unsupervised deep convolutional network. It uses deep features, extracted from encoder-decoder network, and spatial frequency to measure activity level. Experimental results demonstrate that the proposed method achieves the state-of-art fusion performance compared to 16 existing fusion methods in objective and subjective assessment.

Our code and data can be found at \url{https://github.com/Keep-Passion/SESF-Fuse}.

The remainder of this paper is organized as follows: In Section II, we provide a brief review of related works. In Section III, the proposed fusion method is described in detail. The experimental results are shown in Section IV. We conclude the paper in section V.

\begin{figure*}[htb]
\centering
\includegraphics[width=\textwidth]{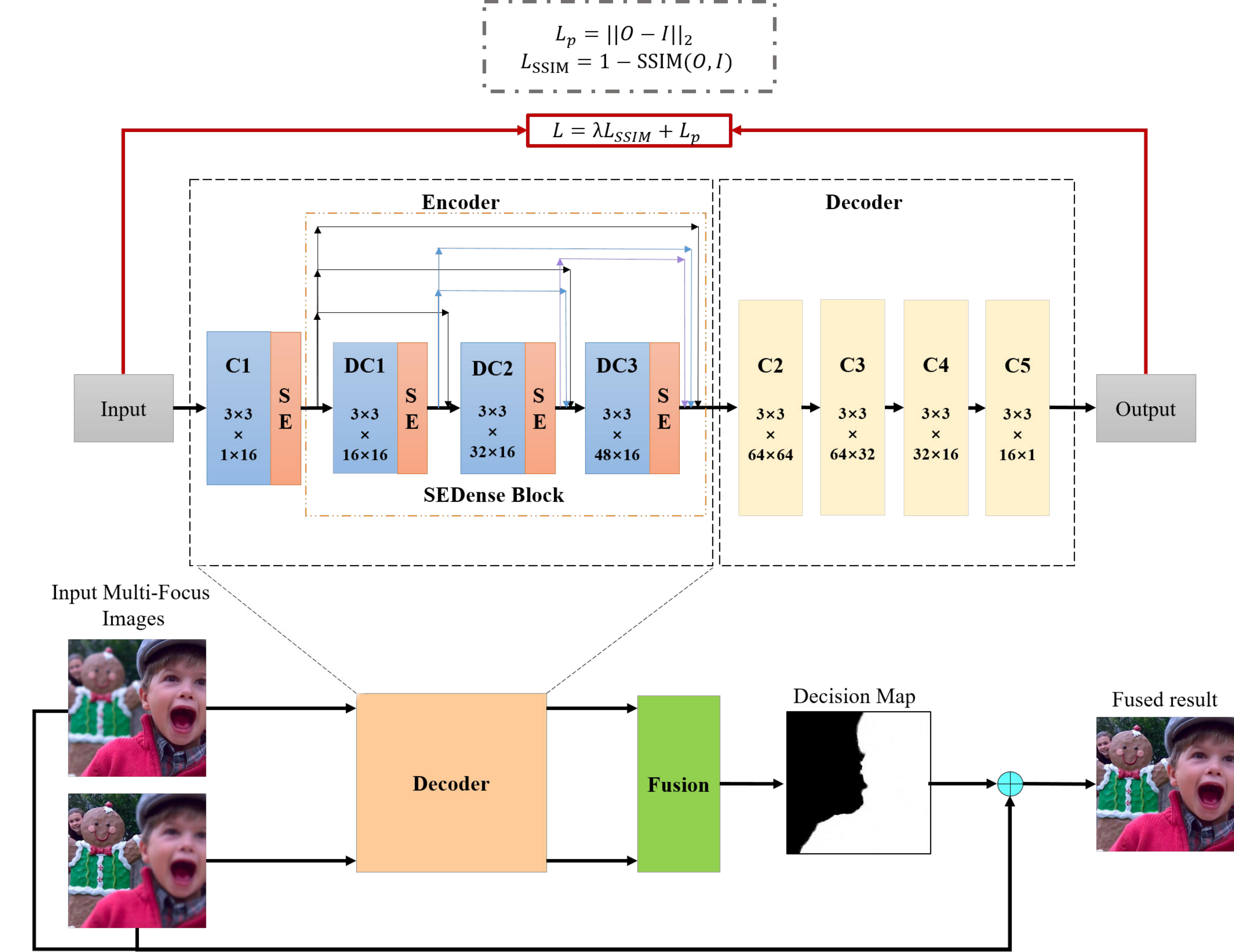}
\caption{The schematic diagram of proposed algorithm.}
\label{fig1}
\end{figure*}

\section{Related work}
\label{sec:realted work}
In the past decades, various image fusion methods have been presented which could be classified into two groups: transform domain methods and spatial domain methods~\cite{stathaki2011image}. The most classical transform domain fusion methods are based on multi-scale transform (MST) theories, such as laplacian pyramid (LP)~\cite{Burt_1983_TOC}, and ratio of low-pass pyramid (RP)~\cite{TOET1989245}, and wavelet-based ones like discrete wavelet transform (DWT)~\cite{LI1995235}, and dual-tree complex wavelet transform (DTCWT)~\cite{LEWIS2007119}, and curvelet transform (CVT)~\cite{NENCINI2007143}, and nonsubsampled contourlet transform (NSCT)~\cite{ZHANG20091334}, and the sparse representation (SR)~\cite{Yang_2010_TIM}, and image matting based (IMF)~\cite{LI2013147}. The key point behind these methods is that the activity level of source images can be measured by the decomposed coefficients in a selected transform domain. Obviously, the selection of transform domain plays a crucial role in these methods. 

Spatial domain fusion methods measure activity level based on gradient information. Early spatial domain fusion methods used manually fixed size block strategy to calculate activity level, spatial frequency for example~\cite{LI2001169}, which usually causes undesirable artifacts. Many improved versions have been proposed on this topic, such as the adaptive block based method~\cite{ASLANTAS20108861} using differential evolution algorithm to obtain a fixed optimal block size. Recently, some pixel-based spatial domain methods based on gradient information have been proposed, such as the guided filtering (GF)-based one~\cite{Li_2013_TIP}, the multi-scale weighted gradient (MWG)-based one~\cite{ZHOU201460} and the dense SIFT (DSIFT)-based one~\cite{LIU2015139}.

With a span of last 5 years, deep convolutional neural network (CNN) has achieved great success in image processing. Some works tried to measure the activity level by high-capacity deep convolutional model. ~\cite{LIU2017191} first applied convolutional neural network to multi-focus image fusion. ~\cite{Prabhakar_2017_ICCV} performed a CNN-based unsupervised approach for exposure fusion problem, which is so called DeepFuse. ~\cite{Li_2019_TIP} presented DenseFuse to fuse infrared and visible images, which used encoder-decoder unsupervised strategy to obtain useful features and fused them by L1-norm. Inspired by DeepFuse, we also train our network in unsupervised encoder-decoder manner. Moreover, we apply spatial frequency as fusing rule to obtain activity level and decision map of source images, which is in accord with the key assumption that only the objects within the depth-of-field have sharp appearance.

\section{Method}
\label{sec:method}

\subsection{Overview of Proposed Method}
\label{sec: overview of the proposed method}
The schematic diagram of our algorithm is shown in Figure \ref{fig1}. We train an auto-encoder network to extract highly dimensional feature in training phase. Then we calculate the activity level using those deep features at fusion layer in inference phase. Finally, we obtain the decision map to fuse two multi-focus source images. The algorithm presented here only aims to fuse two source images. However, to deal with more than two multi-focus images, it can be straightforwardly fuse them one by one in series.

\subsection{Extraction of Deep Feature}
\label{sec: extraction of deep feature}
By getting inspiration from DenseFuse~\cite{Li_2019_TIP}, we only use encoder and decoder to reconstruct the input image and discard fusion operation in training phase. After the encoder and decoder parameters are fixed, we use spatial frequency to calculate the activity level from deep features which are obtained from encoder.

As shown in Figure \ref{fig1}, the encoder consists of two parts(C1 and SEDense Block). C1 is a $3 \times 3$ convolution layer in encoder network. DC1, DC2 and DC3 are $3 \times 3$ convolution layers in SEDense block and the output of each layer is connected to every other layer by cascade operation. In order to reconstruct image precisely, there are no pooling layer in the network. Squeeze and Excitation (SE) block can enhance spatial encoding by adaptively re-calibrating channel-wise feature responses~\cite{Hu_2018_CVPR}, the influence of this structure is shown at the experiment. The decoder consists of C2, C3, C4 and C5, which will be utilized to reconstruct the input image. We minimize the loss function $L$, which combines pixel loss $L_p$ and structural similarity (SSIM) loss $L_{ssim}$, to train our encoder and decoder. $\lambda$ is a constant weight to normalize two loss.

\begin{equation}
\label{loss_formula}
L = \lambda L_{ssim} + L_p
\end{equation}

The pixel loss $L_p$ calculates Euclidean distance between the output($O$) and the input($I$).
\begin{equation}
\label{Lp_formula}
L_p = ||O - I||_2
\end{equation}

The SSIM loss $L_{ssim}$ calculates structural differences between $O$ and $I$. Where $SSIM$ represents to structural similarity operation ~\cite{wang2004image}.

\begin{equation}
\label{L_ssim_formula}
L_{ssim} = 1 - SSIM(O, I)
\end{equation}

\subsection{Detailed Fusion Strategy}
\label{sec: detailed fusion strategy}
\begin{figure}[h]
\centering
\includegraphics[width=\linewidth]{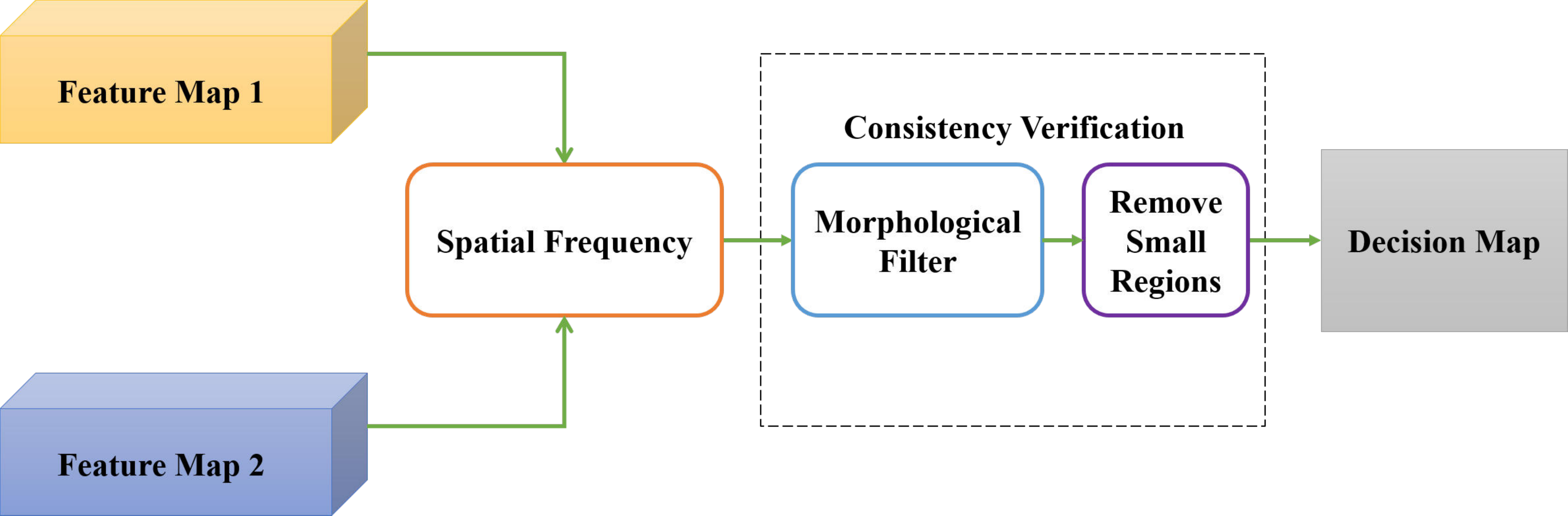}
\caption{The detailed fusion strategy.}
\label{fig2}
\end{figure}

The detailed fusion strategy is shown in Figure \ref{fig2}. We utilize spatial frequency to calculate initial decision map and apply some commonly used consistency verification methods to remove small errors. Finally, we obtain the decision map to fuse two multi-focus source images. 

\subsubsection{Spatial Frequency Calculation using Deep Features}
\label{sec: spatial frequency calculation using deep features}
Different from L1-norm in DenseFuse, we use feature gradient instead of feature intensity to calculate activity level. Specifically, we apply spatial frequency to handle this task using deep features. 

In this paper, the encoder provides high dimensional deep feature for each pixel in an image. However, the original spatial frequency is calculated on gray image with single channel. Thus, for deep features, we modify the spatial frequency calculation method. Let $F$ represents the deep features driven from encoder block. $F_{(x, y)}$ represents one feature vector, $(x, y)$ refers to the coordinates of these vectors in image. We calculate its spatial frequency using the formulas below, where $RF$ and $CF$ are the row and column vector frequency, respectively.

\begin{small}
\begin{equation}
\label{RF}
RF_{(x,y)}=\sqrt{\sum_{a=-r}^{r}\sum_{b=-r}^{r}[F_{(x+a,y+b)}-F_{(x+a, y+b-1)}]^2}
\end{equation}
\end{small}

\begin{small}
\begin{equation}
\label{CF}
CF_{(x,y)}=\sqrt{\sum_{a=-r}^{r}\sum_{b=-r}^{r}[F_{(x+a,y+b)}-F_{(x+a-1, y+b)}]^2}
\end{equation}
\end{small}

\begin{equation}
\label{SF}
SF_{(x,y)} = \sqrt{\frac{{(CF_{(x,y)})}^2+{(RF_{(x,y)})}^2}{(2r+1)^2}}
\end{equation}

Where $r$ is radius of kernel. The original spatial frequency is a block-based method, while it is pixel-based in our method. Besides, we apply 'same' padding strategy at the border of feature maps.

Thus, we can compare the spatial frequencies of two corresponding $SF1$ and $SF2$, where $k$ in $SFk$ is the index of source image. Then we can get the initial decision map (D) with Eq\ref{D}.

\begin{equation}
\label{D}
D_{(x,y)} = 
\begin{cases}
1, & \text{if $SF1_{(x,y)} \ge SF2_{(x,y)}$} \\
0, & \text{otherwise}
\end{cases}
\end{equation}

\begin{figure*}[htb]
\centering
\includegraphics[width=\textwidth]{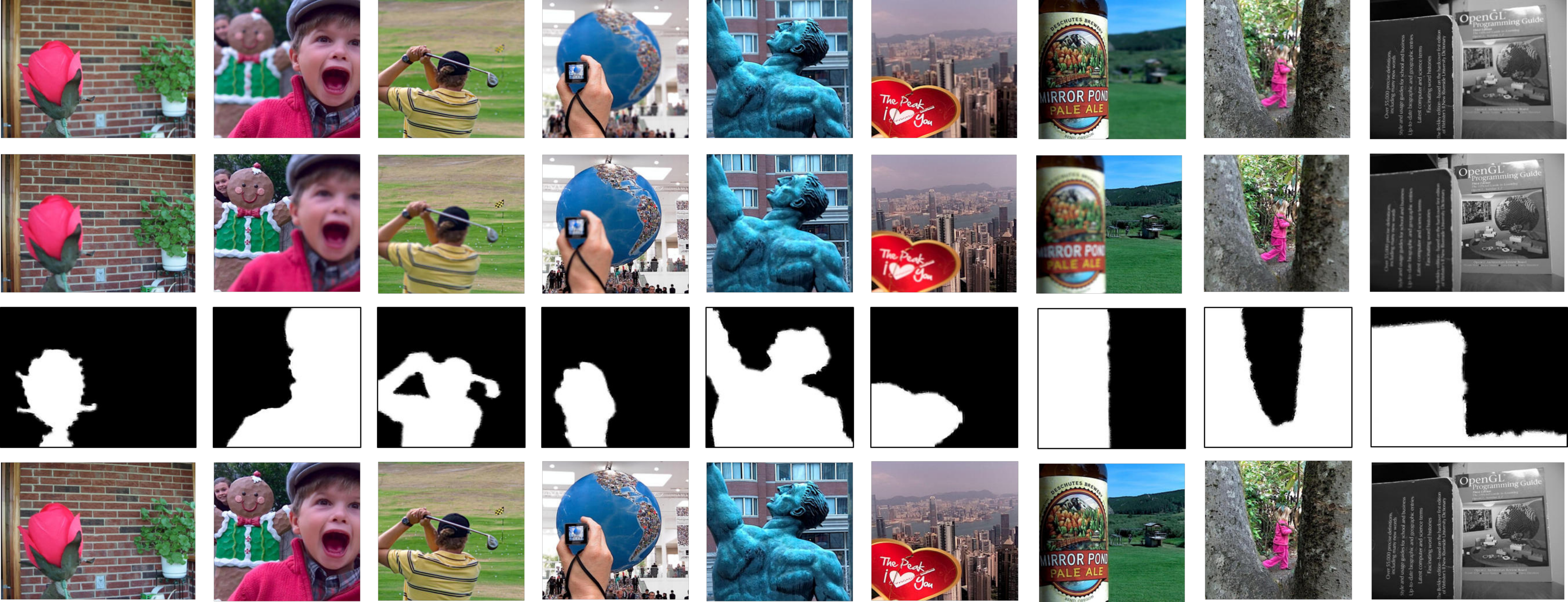}
\caption{Visualization of fused results. The first row is near focused source image and the second row is far focused source image. The third row is decision map of our method and the final row is fused result.}
\label{fig3}
\end{figure*}

\subsubsection{Consistency Verification}
\label{sec: consistency verification}
There may be some small lines or burrs in the connection portions, and some adjacent regions may be disconnected by the inappropriate decisions. Thus, alternating opening and closing operators with a small disk structuring element~\cite{DE20061278} is applied to process the decision map. In this way, the small lines or burrs could be eliminated, the connection portions of the focused regions could be smoothed, and the adjacent regions would be combined as a whole region. We found that, when the radius of the disk structuring element equals to spatial frequency kernel radius, the small lines or burrs could be well detected and the adjacent regions could be connected right. Beside, we apply the small region removal strategy, which is same with ~\cite{LIU2017191}. Specially, we reverse the region which is smaller than an area threshold. In this paper, the threshold is usually set to $0.01 \times H \times W$, where H and W are the height and width of source image, respectively.

Generally, there are some undesirable artifacts around the boundaries between focused and defocused regions. Similar to ~\cite{NEJATI201572}, we utilize an efficient edge-preserving filter, guided filter~\cite{6319316}, to improve the quality of initial decision map, which can transfer the structural information of a guidance image into the filtering result of the input image. The initial fused image is employed as the guidance image to guide the filtering of initial decision map. In this work, we experimentally set local window radius $r$ to 4 and the regularization parameter $\varepsilon$ to 0.1 in guided filter algorithm.

\subsubsection{Fusion}
\label{sec: fusion}
Finally, by using the obtained decision map $D$, we calculate the fused result $F$ with the following pixel-wise weighted-average rule. The input images are denoted as $Imgk$ which are pre-registered, where $k$ represents the index of source images. The representative visualization of fused images are shown in Figure \ref{fig3}.
\begin{equation}
\label{fusion_formula}
F{_{(x, y)}} = D_{(x, y)}Img1_{(x, y)} + (1 - D_{(x, y)})Img2_{(x, y)}
\end{equation}

\section{Experiments}
\label{sec: experimentas}
\subsection{Experimental Settings}
\label{sec: experimental settings}
In our experiment, we use 38 pairs of multi-focus images as testing set for evaluation, which are publicly available online~\cite{NEJATI201572,savic2012multifocus}.

Due to the unsupervised strategy, we first train the encoder-decoder network using MS-COCO~\cite{COCO}. In this phase, about 82783 images are utilized as training set, 40504 images are used to validate the reconstruction ability in every iteration. All of them are resized to $256 \times 256$ and transformed to gray scale images. Learning rate is set as $1 \times 10^{-4}$ and then decrease by a factor of 0.8 at every two epoch. We set $\lambda=3$ which is same with DenseFuse~\cite{Li_2019_TIP} and optimize the objective function with respect to the weights at all network layer by Adam~\cite{kingma2015adam}. The batch size and epochs are 48 and 30, respectively. And then we used acquired parameters to perform SF fusion on the testing set above. 

Our implementation of this algorithm is derived from the publicly available Pytorch framework~\cite{pytorch}. The network’s training and testing are performed on a system using 4 NVIDIA 1080Ti GPU with 44GB memory.

\begin{figure*}[htb]
\centering
\includegraphics[width=\textwidth]{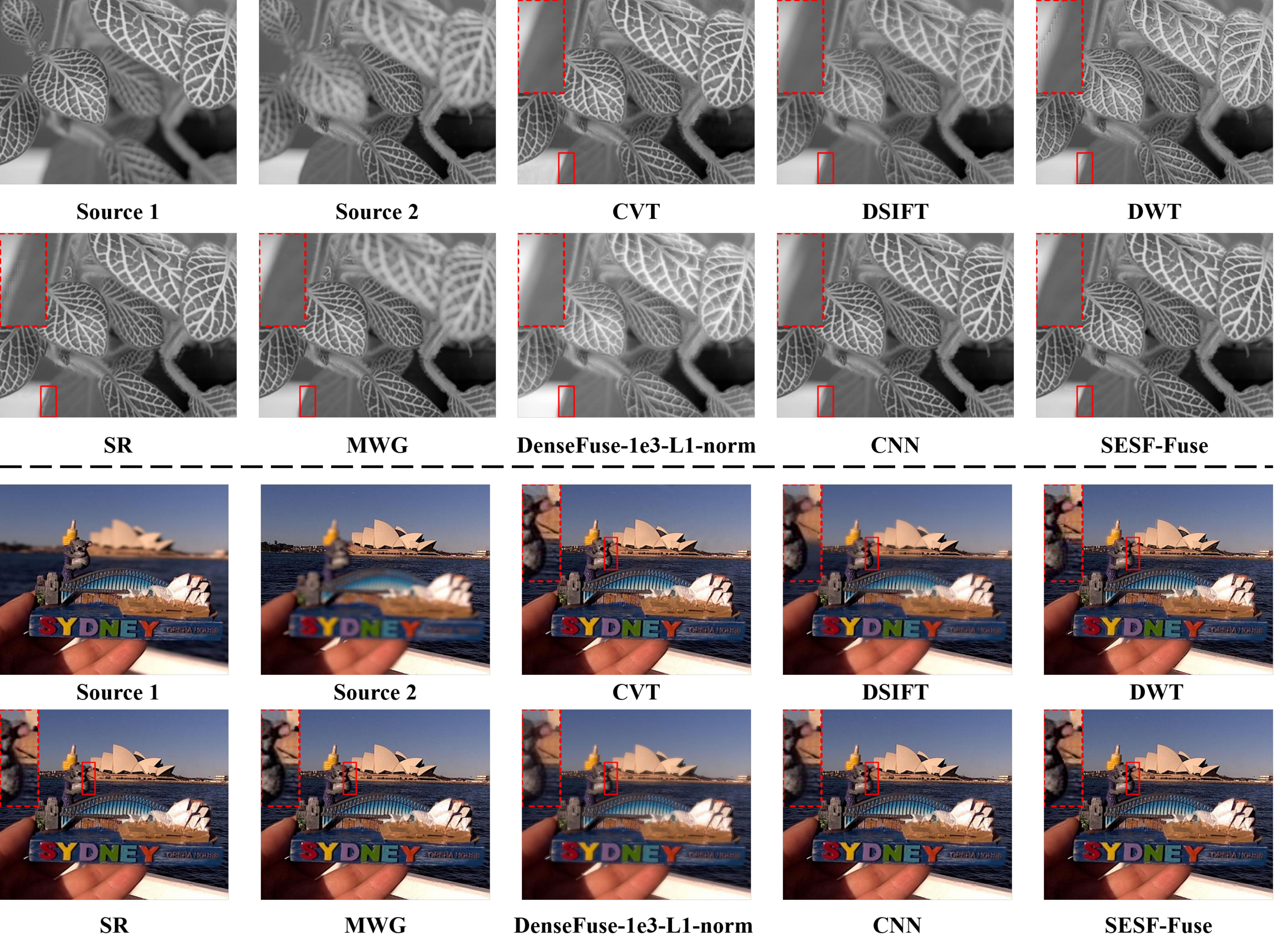}
\caption{Visualization of different 'leaf' and 'Sydney Opera House' fused results.}
\label{fig4}
\end{figure*}

\subsection{Objective Image Fusion Quality Metrics}
\label{sec: objective image fusion quality metrics}
The proposed fusion method is compared with 16 representative image fusion methods, which are the laplacian pyramid (LP)-based one~\cite{Burt_1983_TOC}, the ratio of low-pass pyramid (RP)-based one~\cite{TOET1989245}, the nonsubsampled contourlet transform (NSCT)-based one~\cite{ZHANG20091334}, the discrete wavelet transform (DWT)-based one~\cite{LI1995235}, dual-tree complex wavelet transform (DTCWT)-based one~\cite{LEWIS2007119}, the sparse representation (SR)-based one~\cite{Yang_2010_TIM}, the curvelet transform (CVT)-based one~\cite{NENCINI2007143}, the guided filtering (GF)-based one~\cite{Li_2013_TIP}, the multi-scale weighted gradient (MWG)-based one~\cite{ZHOU201460}, the dense SIFT (DSIFT)-based one~\cite{LIU2015139}, the spatial frequency(SF)-based one~\cite{LI2001169}, the the FocusStack~\cite{FocusStack}, the Image Matting Fusion(IMF)~\cite{LI2013147}, the DeepFuse~\cite{Prabhakar_2017_ICCV}, the DenseFuse (both add and L1-norm fusion strategy)~\cite{Li_2019_TIP} and the CNN-Fuse~\cite{LIU2017191}. In addition, GF,  IMF are driven from~\cite{xudongkangweebly} and NSCT, CVT, DWT, DTCWT, LP, RP, SR and CNN-Fuse from~\cite{LiuyuCodes}.

In order to access the fusion performance of different methods objectively, we adopt three fusion quality metrics, such as $Q_g$~\cite{840225}, $Q_m$~\cite{4697288} and $Q_{cb}$~\cite{CHEN20091421}. For each of the above three metrics, a larger value indicates a better fusion performance. A good comprehensive survey of quality metrics can be found in the article~\cite{Liu_2012_TIP}. For fair comparison, we use default parameters given in the related publications for these metrics and all codes are driven from ~\cite{imageFusionMetrics}.

\begin{figure*}[htb]
\centering
\includegraphics[width=\textwidth]{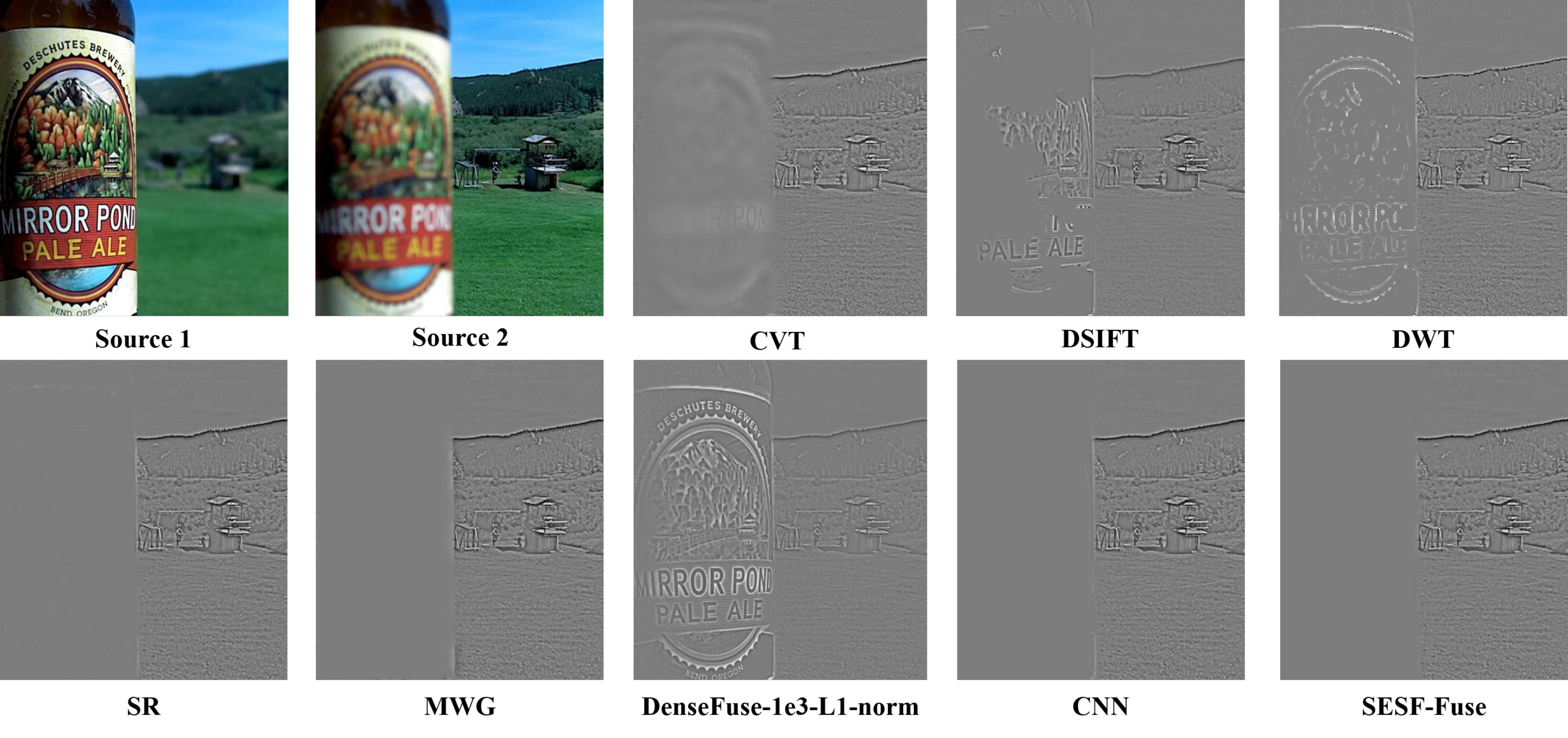}
\caption{The difference images for each 'beer' fused results}
\label{fig5}
\end{figure*}

\subsection{Ablation Experiments}
\label{sec: ablation experiments}
We first evaluate our methods with different settings to verify our methods. We pick up seven fusion modes to explore the usage of deep features, such as max, abs-max, average, L1-norm, sf, se\_sf\_dm, and dense\_sf\_dm. DenseFuse~\cite{Li_2019_TIP} investigated add and L1-norm fusion strategy and draw out the conclusion that L1-norm of deep feature could be used to fuse infrared-visible images. They utilized feature intensity to calculate activity level. We found that feature gradient (calculated by spatial frequency) is suited to multi-focus fusion task. Table \ref{table1} shows mean average score with different methods. The bold value denotes the best performance among all fusion modes. The digits within a parenthesis indicates the number of results on which corresponding methods obtain the first place. Se\_sf outperforms abs-max, max, average, l1\_norm fusion modes in metric evaluation. In addition, even though the deep learning has promising representative ability, it can not recover the image perfectly. Thus if we use sf to fuse the deep features and input to decoder and draw out result, the fused result could not completely recover every detail of in-focus region. Therefore, we propose to use deep features to calculate the decision map and fuse the original images. As shown in experiment results, the performance of se\_sf\_dm defeats the se\_sf's. Besides, we conduct an experiment to verify the influence of SE architecture~\cite{Hu_2018_CVPR}, we have found that the average scores of se\_sf\_dm in $Q_g$ and $Q_m$ is higher than dense\_sf\_dm and the first place number of se\_sf\_dm is the highest result. We assume that squeeze-and-excitation structure could dynamically recalibrate feature which shows robust result.

\begin{table}[htb]
\caption{Ablation experiments with different settings.}
\label{table1}
\begin{tabular}{cccc}
\hline
\textbf{Methods}    & $Q_g$               & $Q_m$               & $Q_{cb}$           \\ \hline
\textbf{se\_absmax}          & 0.5204(0)           & 2.4880(0)           & 0.6019(0)           \\
\textbf{se\_average}         & 0.5033(0)           & 2.4835(0)           & 0.5963(0)           \\
\textbf{se\_l1\_norm}        & 0.5124(0)           & 2.4961(0)           & 0.6020(0)           \\
\textbf{se\_max}             & 0.5059(0)           & 2.4851(0)           & 0.5980(0)           \\
\textbf{se\_sf}              & 0.6885(0)           & 2.7216(2)           & 0.7526(0)           \\
\textbf{se\_sf\_dm} & \textbf{0.7105(25)} & \textbf{2.8886(16)} & 0.7848(19)          \\
\textbf{dense\_sf\_dm}       & 0.7103(13)          & 2.8872(20)          & \textbf{0.7852(19)} \\ \hline
\end{tabular}
\end{table}

\subsection{Comparison with other fusion methods}
\label{sec: comparisons with other fusion methods}
We first compare the performance of different fusion methods based on visual perception. For this purpose, four examples in two manners are mainly provided to exhibit the difference among different methods.  

In Figure \ref{fig4}, we visualize two fused examples, such as 'leaf' and 'Sydney Opera House' image pairs and their fused results. In each image, a region around the boundary between focused and defocused parts is magnified and shown in the higher left corner. In 'leaf' result, we can see that the border of leaf with different methods. The DWT shows 'serrated' shape and the CVT, DSIFT, SR, DenseFuse, CNN show undesirable artifacts. Besides, for DWT and DenseFuse, the luminance of leaf at right higher corner shows an abnormal increase. And the same region in MWG is out-of-focused, which means that the method can not well detect the focused regions. In 'Sydney Opera House' result, the ear of Koala located at the border between focused and defocused parts, as we can see that all methods show smooth and blurred results except SESF-Fuse.

\begin{figure*}[htb]
\centering
\includegraphics[width=\textwidth]{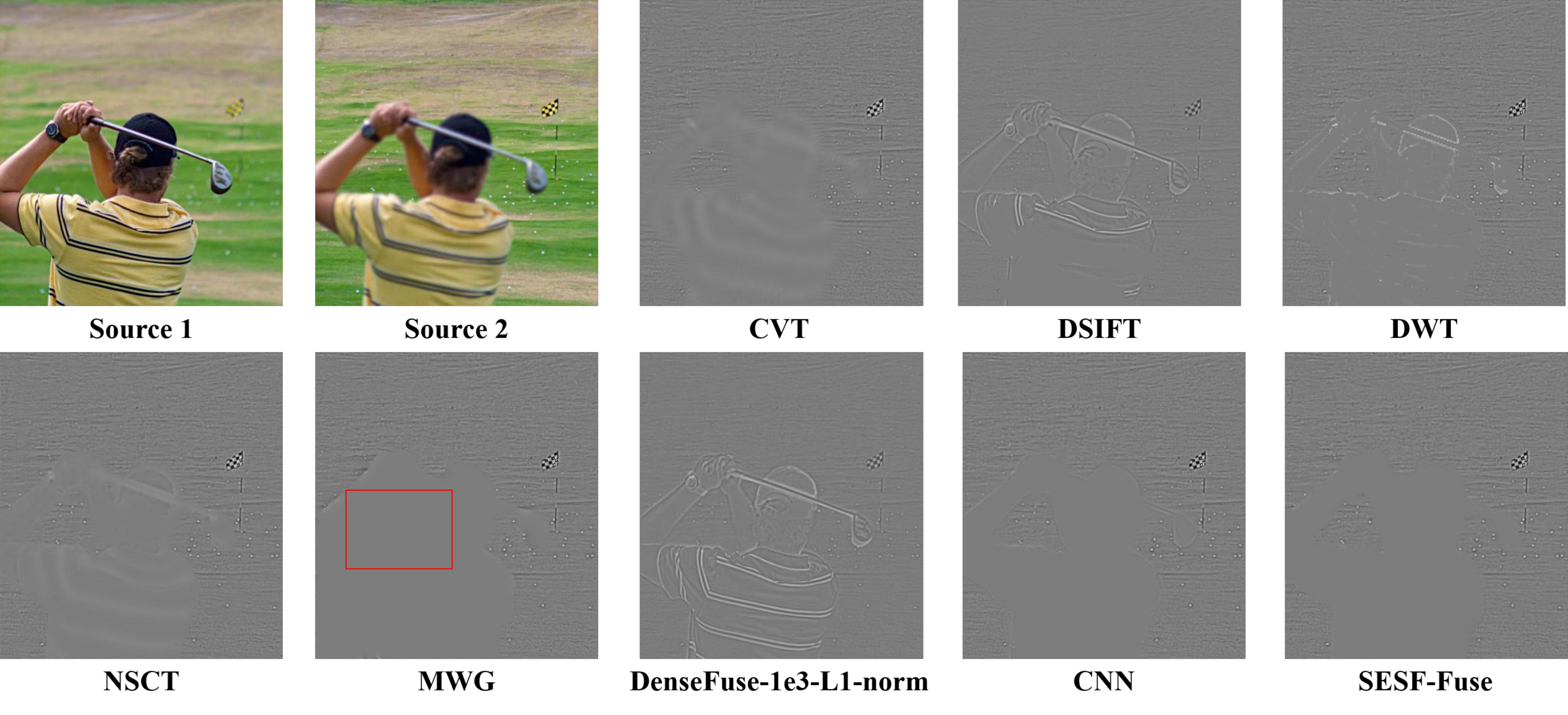}
\caption{The difference images for each 'golf' fused results}
\label{fig6}
\end{figure*}

To have a better comparison, Figure \ref{fig5} and Figure \ref{fig6} show the difference images obtained by subtracting the first source image from each fused image, and the values of each difference image are normalized to the range of 0 to 1. If the near focused region is completely detected, the difference image will not show any information of that. In Figure \ref{fig5}, it is beer bottle. Therefore, the CVT, DSIFT, DWT and DenseFuse-1e3-L1-Norm can not perfectly detect the focused region. The SR, MWG and CNN perform well except the region at the border of bottle, because we still can see the contour of near focused region. Besides, our SESF-Fuse performs well in both center or border region of near focused regions. In Figure \ref{fig6}, the near focus region is the man. Same with the observation above, the CVT, DSIFT, DWT, NSCT, DenseFuse can not perfectly detect the focused region. The MWG and CNN perform well except that the region at the border of the person. Besides, for MWG, the region surrounded by arms is actually far focused region, MWG can not correctly detect here.

\begin{table*}[htb]
\centering
\caption{Comparison with other fusion methods.}
\label{table2}
\begin{tabular*}{\hsize}{@{}@{\extracolsep{\fill}}ccccccc@{}}
\hline
\textbf{Metrics} & \textbf{DeepFuse} & \textbf{FocusStack} & \textbf{SF}        & \textbf{DenseFuse\_1e3\_add} & \textbf{DSIFT}      & \textbf{DenseFuse\_1e3\_l1} \\ \hline
$Q_g$      & 0.4269(0)  & 0.4709(0)    & 0.5115(0) & 0.5190(0)             & 0.5267(0)  & 0.5283(0)            \\
$Q_m$      & 2.4618(0)  & 2.8510(0)    & 2.8512(0) & 2.8530(0)             & 2.8725(0)  & 2.8561(0)            \\
$Q_{cb}$     & 0.5651(0)  & 0.6330(0)    & 0.6024(0) & 0.6008(0)             & 0.6067(0)  & 0.5972(0)            \\ \hline
\textbf{Metrics} & \textbf{GF}         & \textbf{CVT}          & \textbf{DWT}       & \textbf{IMF}                   & \textbf{RP}         & \textbf{DTCWT}                \\ \hline
$Q_g$      & 0.5631(0)  & 0.6187(0)    & 0.6222(0) & 0.6324(2)             & 0.6478(0)  & 0.6529(0)            \\
$Q_m$      & 2.8506(0)  & 2.9563(0)    & 2.9465(1) & 2.8844(0)             & 2.9460(0)  & 2.9583(0)            \\
$Q_{cb}$     & 0.7008(3)  & 0.6908(0)    & 0.6712(0) & 0.7362(4)             & 0.7101(0)  & 0.7126(0)            \\ \hline
\textbf{Metrics} & \textbf{NSCT}       & \textbf{SR}           & \textbf{LP}        & \textbf{MWG}                   & \textbf{CNN-Fuse}  & \textbf{SESF-fuse}  \\ \hline
$Q_g$      & 0.6587(0)  & 0.6686(0)    & 0.6731(0) & 0.6998(0)             & 0.7102(16) & \textbf{0.7105(20)}  \\
$Q_m$      & 2.9592(0)  & 2.9630(2)    & 2.9642(8) & 2.9615(6)             & 2.9654(7)  & \textbf{2.8886(14)}  \\
$Q_{cb}$     & 0.7169(0)  & 0.7335(0)    & 0.7352(0) & 0.7764(2)             & 0.7839(9)  & \textbf{0.7848(20)}  \\ \hline
\end{tabular*}
\end{table*}

Table \ref{table2} lists the objective performance of different fusion methods using the above three metrics. We can see that the CNN-based method and the proposed method clearly beat the other 15 methods on the average score of $Q_g$ and $Q_{cb}$ fusion metrics. For $Q_g$ metric, CNN-Fuse and SESF-Fuse achieve comparable performance. However, CNN-Fuse is a supervised method which needs to generate synthetic images with different blurred levels to train a two-class image classification network. By contrast, our network only needs to train an unsupervised model which doesn't need to generate synthetic image data. And for $Q_m$ metric, the average score of SESF-Fuse is smaller than LP, however, the first place number of proposed method achieves the highest value which means it is more robust than other methods.

Considering the above comparisons on subjective visual quality and objective evaluation metrics together, our proposed SESF-Fuse-based fusion method can generally outperform other methods, leading to state-of-the-art performance in multi-focus image fusion.

\section{Conclusion}
\label{sec:conclusion}
In this work, we propose an unsupervised deep learning model to address multi-focus image fusion problem. First, we train an encoder-decoder network in unsupervised manner to acquire deep feature of input images. And then we utilize these features and spatial frequency to calculate activity level and decision map to perform image fusion. Experimental results demonstrate that the proposed method achieves the promising fusion performance compared to existing fusion methods in objective and subjective assessment. This paper demonstrate the viability of combination of unsupervised learning and traditional image processing algorithm. Our team will promote this research in subsequent work. Besides, we believe that same strategy could be applied to other image fusion tasks, such as multi-exposure fusion, infrared-visible fusion and medical image fusion.

\section{Acknowledgments}
\label{sec:acknowledgement}
The authors acknowledge the financial support from the National Key Research and Development Program of China (No. 2016YFB0700500), and the National Science Foundation of China (No. 61572075, No. 61702036, No. 61873299, No. 51574027), and Key Research Plan of Hainan Province (No. ZDYF2018139).

\bibliographystyle{aaai}
\bibliography{main}

\end{document}